\documentclass[runningheads]{llncs}
\usepackage[T1]{fontenc}
\usepackage{graphicx}
\usepackage{hyperref}
\usepackage{color}

\usepackage{DeclareUnicodeCharacter}
\usepackage{listings}
\usepackage{xcolor}
\usepackage{booktabs}
\usepackage{multirow}
\usepackage{makecell}
\usepackage[caption=false]{subfig}
\usepackage{makecell}

\definecolor{codegreen}{rgb}{0,0.6,0}
\definecolor{codegray}{rgb}{0.5,0.5,0.5}
\definecolor{codepurple}{rgb}{0.58,0,0.82}
\definecolor{backcolour}{rgb}{0.95,0.95,0.92}

\lstdefinestyle{mystyle}{
    backgroundcolor=\color{backcolour},   
    commentstyle=\color{codegreen},
    keywordstyle=\color{magenta},
    numberstyle=\tiny\color{codegray},
    stringstyle=\color{codepurple},
    basicstyle=\ttfamily\footnotesize,
    breakatwhitespace=false,
    escapeinside={\%*}{*)},
    breaklines=true,                 
    captionpos=b,                    
    keepspaces=true,                 
    numbers=left,                    
    numbersep=10pt,                  
    showspaces=false,                
    showstringspaces=false,
    showtabs=false,                  
    tabsize=2
}

\usepackage[most]{tcolorbox}
\newtcolorbox{promptbox}[1]{
    breakable,
    colback=gray!5,
    colframe=gray!75,
    boxrule=0.5pt,
    arc=3pt,
    left=4pt,
    right=4pt,
    top=4pt,
    bottom=4pt,
    fonttitle=\bfseries,
    title=#1,
    fontupper=\small
}

\begin{document}
\title{Are you Talking Logic to Me? Assessing Language Models Syllogistic Reasoning Capabilities}
\titlerunning{Are you Talking Logic to Me?}
%
\author{Hanna Abi Akl\inst{1,2}\orcidID{0000-0001-9829-7401} \and
Fabien Gandon\inst{1}\orcidID{0000-0003-0543-1232} \and
Catherine Faron\inst{1}\orcidID{0000-0001-5959-5561} \and
Pierre Monnin\inst{1}\orcidID{0000-0002-2017-8426}}
\authorrunning{H. Abi Akl et al.}
%
\institute{Université Côte d’Azur, Inria, CNRS, I3S, Sophia Antipolis, France \and
Data ScienceTech Institute, Paris, France
\\
\email{\{hanna.abi-akl,fabien.gandon,pierre.monnin\}@inria.fr, faron@i3s.unice.fr}}
\maketitle              
\begin{abstract}
Language models (LMs) struggle with logical tasks like reasoning on syllogisms. 
It has been shown that Knowledge Representation (KR) plays a crucial role in expressing input information to help models solve tasks. 
This observation motivates our study of the impact of different formal KR notations on syllogistic reasoning by extending the FOLIO and P-FOLIO datasets. Our experiments on Small Language Models (SLMs) in Supervised Fine-Tuning (SFT) and Zero-Shot (ZS) settings show that the choice of input notation can yield performances competitive with natural language while enabling faster inference. We also propose a syllogistic categorization method (SEF) and use it to enrich ZS prompts with logical definitions, which boost reasoning in small models. 
We open-source our framework, Common Logic Grammar Construction (CLGC), as the first Python library for automatically generating syllogisms in KR notations and defining their SEF categories.

\end{abstract}
\section{Introduction}
\label{sec:introduction}

Advances in Artificial Intelligence (AI) techniques have seen Language Models (LMs) hit new heights in solving tasks of increasing complexity. However, logic tasks like planning or puzzle solving still represent a challenging frontier.
In this paper, we focus on generalized syllogistic reasoning covering multiple forms of syllogisms with new datasets and LM-based techniques to increase model robustness and consistency~\cite{bertolazzi2024systematic,dasgupta2022language,eisape2024systematic,ozeki2024exploring}. 
Past research has shown that the complexity of this task starts with the dataset, where current datasets suffer from shortcomings like insufficient human involvement, size, diversity, or generalization~\cite{han2024folio,han2024p,saparov2022language,tafjord2021proofwriter,kwon2025logicqa,chen2025justlogic}. 
Moreover, recent neuro-symbolic approaches have used symbolic data representations like First-Order Logic (FOL) due to their interpretable and unambiguous nature as opposed to Natural Language (NL). 
These approaches have shown promising results in boosting model performance on reasoning tasks, most notably small and medium LMs~\cite{quan2024verification,wysocka2025syllobio}. 
However, to the best of our knowledge, syllogistic reasoning is currently evaluated only in NL and FOL notations, 
whereas other shades of Knowledge Representation (KR) languages with varying levels of abstraction between natural and symbolic exist to represent logical data.
Additional metadata about the reasoning task, e.g. categories of syllogisms, could also be considered to support LM reasoning.
Varying the input data representation and studying its effects on logical reasoning thus appears as a necessary challenge to evaluate the limits of LMs in scenarios requiring complex abstract thinking~\cite{wang2025comprehensive}.
These observations motivate our work, where we aim to address the following Research Questions (RQs):
\begin{description}
    \item[RQ1.] How do different input formal notations impact LM syllogistic reasoning?
    \item[RQ2.] How does grouping syllogisms in pre-defined categories and prompting LMs with this information affect their reasoning abilities?
\end{description}

We propose to answer these RQs by providing a protocol comprising automatic formal notation generation, syllogism category definition and classification, and evaluation of Small Language Models (SLMs) on syllogism datasets in Supervised Fine-Tuning (SFT) and Zero-Shot (ZS) settings. 
We specifically focus on SLMs due to their frugality and performances on similar tasks~\cite{akl:hal-05606250,quan2024verification,wysocka2025syllobio}.
Our contribution to the field is threefold:
\begin{itemize}
    \item We introduce the Common Logic Grammar Construction framework for generating syllogisms in different formal notations as well as categorizing them based on their structure; 
    we open-source it as a Python package.
    \item We release an enriched, extended version of the FOLIO and P-FOLIO reasoning datasets with formal notations for our experiments.
    \item We evaluate the impact of several formal notations on the performance of LM on  syllogistic reasoning and establish different trends based on the abstraction level of the notation.
\end{itemize}
The rest of this paper is organized as follows: we discuss the related work in Section \ref{sec:relatedwork}, present our methodology in Section \ref{sec:clgc}, describe our experimental setup in Section \ref{sec:experimental-setup}, discuss our findings in Section \ref{sec:results} and conclude in Section \ref{sec:conclusion}.

\section{Related Work}
\label{sec:relatedwork}

\subsection{Language Models and Reasoning}
\label{subsec:lm-reasoning}

Logic in AI deals with tasks involving abstraction like syllogistic reasoning~\cite{valentino-etal-2026-semeval}. 
Current neuro-symbolic approaches have used symbolic structures to improve language model reasoning. Advanced prompting techniques with rule templates have shown increases in performance and more control on reasoning bias in syllogisms~\cite{seals2024evaluating,valentino2026mitigating}. Other methods use Chain-Of-Thought (COT) to generate reasoning steps and explanations as a self-correction mechanism for LMs~\cite{xu2024faithful,lyu-etal-2023-faithful}. 
Recently, hybrid architectures have used NL syllogisms and FOL translations with a theorem prover to guide model reasoning~\cite{ranaldi2025improving,kim2025reasoning,maraia2026abstract,hoppe2025investigating}. Our work leverages different input representations as logical basis to improve reasoning. Training SLMs with different input representations has shown promising results compared to NL baselines on syllogisms~\cite{akl:hal-05248053,akl:hal-05606250}.
In that perspective, we use existing yet under-explored KR notation families proposed in the literature, namely: the Common Logic Interchange Format (CLIF) and Conceptual Graph Interchange Format (CGIF)~\cite{sowa2016conceptual,sowa2008conceptual,sowa1992conceptual,sowa2011introduction} from the Common Logic (CL) suite, Tensor Functor Logic Plus (TFLPLUS)~\cite{sommers2017invitation,castro2018programming,manzano2019intermediate} from Plus-Minus Algebra, and CLINGO~\cite{gebser2014clingo} from Answer Set Programming (ASP) languages. To the best of our knowledge, no other method in the literature adopts and evaluates the impact of KR notations on syllogistic reasoning. 

\subsection{LM Experimentation on Syllogisms}
\label{subsec:lm-iterative-experimentation}

The nature of LM experimentation and the need for an efficient iterative approach due to the sensitivity of models to different parameters such as input data justifies the need for standardized frameworks for reasoning tasks. To the best of our knowledge, no such framework exists today to study syllogisms data in different representations. The closest initiatives are solvers in Python\footnote{\url{https://github.com/czrptr/syllogism-solver}, \url{https://github.com/mhtess/syllogism}} and Lua\footnote{\url{https://pypi.org/project/syllogistic/}} but they operate with syllogisms in NL. To address this gap, we propose the Common Logic Grammar Construction framework (CLGC)\footnote{\url{https://pypi.org/project/clgc/}}, a Python package to automatically manipulate syllogisms in different formal notations.

\section{Syllogistic Reasoning: From Formal Notation Generation to LM Evaluation}
\label{sec:clgc}

\subsection{CLGC Formal Notation Generation}
\label{subsec:notation-generation}

\subsubsection{Notation Selection Criteria. }
\label{subsubsec:selection-criteria}

We define the following criteria:
\begin{description}
    \item[Verbosity.] Our set should include compact (i.e. with minimal syntactic variations) and verbose (i.e. with many syntactic variations) notations.
    Greater verbosity leaves more room for ambiguity when expressing a syllogism. 
    \item[Frequency.] Frequency is relative to the LMs used in our experimental setup (Section \ref{sec:experimental-setup}). Since LM pre-training is not always fully disclosed, we select notations that are frequently seen  by LMs (e.g. NL) versus notations we assume are much less seen (e.g. CLIF), to the best of our estimation.
    \item[Abstractness.] Abstractness defines the syntactic complexity of a notation, ranging from natural (i.e. NL) to symbolic, mathematical notations. 
    \item[Finiteness.] A hard requirement is for notations to have a finite vocabulary and definable grammar. 
\end{description}

\subsubsection{Generation Process. }
\label{subsec:generation-process}

To generate a formal notation, we start from syllogisms in FOL. FOL meets the above defined criteria and has a defined grammar in Backus-Naur Form (BNF). We reuse the algorithm previously introduced in~\cite{akl:hal-05248053} to automatically generate a formal notation from FOL. The algorithm consists in first generating the Abstract Syntax Tree (AST) from the FOL BNF grammar and then constructing an equivalent tree in the target notation using the notation's BNF grammar that we manually implement in accordance with the above defined finiteness criteria. From the resulting AST, a parser reconstructs statements in the target notation while applying syntactic rules specific to the notation (e.g. spacing, parentheses). To make the process reproducible, we packaged and open-sourced our pipeline in the Python CLGC\footnote{Named as an ode to John Sowa} library\footnote{\url{https://github.com/HannaAbiAkl/clgc}}.

\subsubsection{Generated Notations.}
\label{subsec:generated-notations}

\begin{figure}
    \centering
    \includegraphics[width=0.8\textwidth]{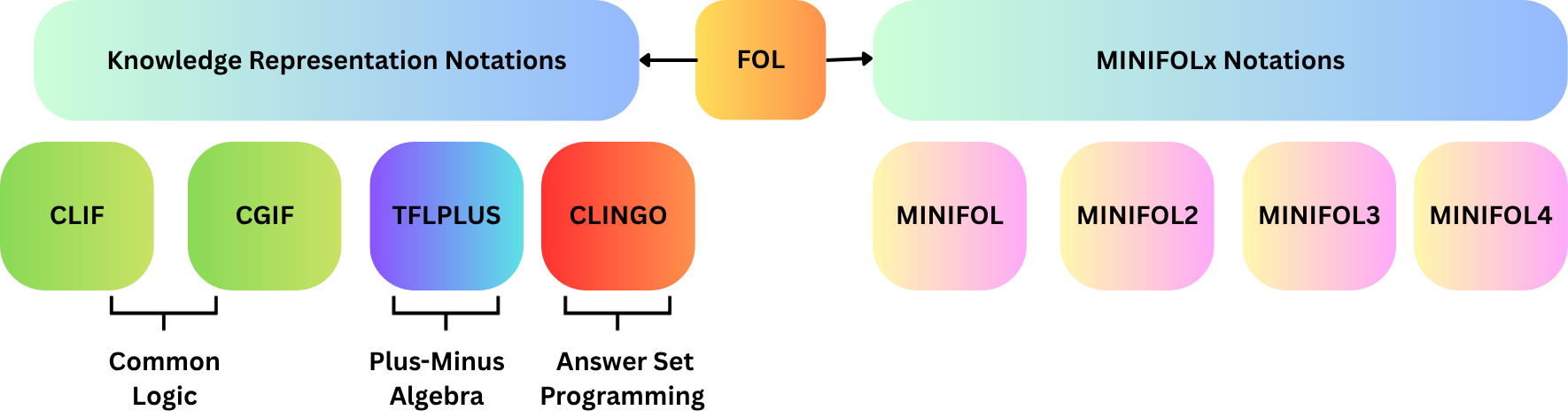}
    \caption{CLGC Notations.} \label{fig:clgc-languages}
\end{figure}

Figure \ref{fig:clgc-languages} shows the current formal notations supported in CLGC. The KR suite contains families of formal notations like CL languages (i.e. CLIF, CGIF), abstract notations like the Plus-Minus Algebra family (i.e. TFLPLUS), and the ASP family (i.e. CLINGO).

The Miniature FOL (MINIFOL\textit{x}) suite is a family of previously unseen lightweight variants of FOL that we introduce as a baseline for studying how minor syntactic variations in notation affect LM syllogistic reasoning performance, thereby enabling to assess sensitivity to the Frequency criterion. 
Precisely, MINIFOL replaces some FOL vocabulary (i.e. symbols) with vocabulary that is more common for LMs (i.e. words).
Here, \textit{x} denotes a specific variant (e.g., MINIFOL\textit{1}). We extend the definition of MINIFOL from~\cite{akl:hal-05606250} to the variants introduced in this work:

\begin{description}
    \item[MINIFOL.] Replaces FOL operators (namely: $\forall$, $\oplus$, $\rightarrow$, $\neg$, $\exists$, $\wedge$, $\vee$) by strings (respectively: all, \^~ , :-, \~~ ,	some, \&, |). 
    \item[MINIFOL2.] Eliminates the $\exists$ quantifier in MINIFOL.
    \item[MINIFOL3.] Replaces the $\neg$ operator with \textit{not} in MINIFOL.
    \item[MINIFOL4.] Replaces the $\wedge$ operator with \textit{,} in MINIFOL.
\end{description}

\subsection{Syllogism Evaluation Framework (SEF)}
\label{subsec:sef}

SEF is a categorization mechanism previously introduced in~\cite{akl:hal-05248053} to classify syllogisms based on their structure in one of the four following categories:
\begin{description}
    \item[Hypothetical.] Syllogism containing an implication
    \item[Disjunctive.] Syllogism containing a disjunction
    \item[Categorical.] Syllogism of exactly 2 premises not belonging to the aforementioned categories
    \item[Complex.] Syllogism not belonging to any other category
\end{description}
Each category is associated with a definition leveraged by CLGC to automatically detect a syllogism's SEF category.
These definitions can also be used to guide syllogistic reasoning. Table \ref{tab:sef-example} illustrates the SEF categories.

\begin{table}
\caption{SEF category examples from P-FOLIO-KR.}\label{tab:sef-example}
\begin{center}
\begin{tabular}{lc}
\toprule
\multicolumn{1}{c}{SEF Category} & \multicolumn{1}{c}{Syllogism example} \\
\midrule
Complex & \makecell{RenamedAs(fortCarillon, fortTiconderoga) \\
Built(pierredeRigauddeVaudreuil, fortCarillon) \\
LocatedIn(fortCarillon, newFrance) \\
\rule{0.5\textwidth}{0.1pt}\\
$\neg$LocatedIn(newFrance, europe)} \\
\midrule
Categorical & \makecell{SportingEvent(olympics) \\
LastSummerOlympics(tokyo) \\
\rule{0.5\textwidth}{0.1pt}\\
MostMedals(unitedStates, tokyo)} \\
\midrule
Hypothetical & \makecell{$\forall$x (Square(x) $\rightarrow$ FourSided(x)) \\
$\forall$x (FourSided(x) $\rightarrow$ Shape(x)) \\
\rule{0.5\textwidth}{0.1pt}\\
$\forall$x (Square(x) $\rightarrow$ Shape(x))} \\
\midrule
Disjunctive & \makecell{ FlyTo(susan, lgaAirport) \\
$\forall$x $\forall$y (FlyFrom(x, y) $\oplus$ FlyTo(x, y)) \\
\rule{0.5\textwidth}{0.1pt}\\
FlyFrom(john, lgaAirport)} \\
\bottomrule
\end{tabular}
\end{center}
\end{table}

\subsection{Evaluating LM Reasoning Capability}
\label{subsec:lm-reasoning-evaluation}

We formalize the task as follows.
A syllogism \textit{S} is a logical problem defined by a set of statements called premises and a conclusion. 
The goal is to verify if the conclusion  logically follows from the premises by labeling the conclusion \textit{True} (i.e. valid), \textit{False} (i.e. invalid), or \textit{Unknown} (i.e. inconclusive).

We conduct our evaluation in SFT and ZS settings. In SFT, we aim to evaluate how well a model can learn a formal notation by training it on syllogisms in that notation and their truth labels. The model is then tasked to predict the truth label of syllogisms formalized in the same notation.
In ZS, we consider two scenarios. \textit{Scenario 1} prompts the model to predict the truth label of a syllogism in a formal notation given the syllogism and the BNF grammar of the notation.
\textit{Scenario 2} adds the SEF category of the syllogism with its definition, and an example to the prompt. The aim of providing this additional information is to facilitate the understanding of logical problems and guide the reasoning process.
Below is the template of each prompt scenario for the FOL notation. 

\begin{promptbox}{Scenario 1}
 You are an expert logician. You are given a syllogism in FOL with premises between <PREMISES></PREMISES> and conclusion between <CONCLUSION></CONCLUSION> tags.
The FOL BNF grammar to understand and reason in the language is given in the <GRAMMAR></GRAMMAR> tags.
\\<GRAMMAR>...</GRAMMAR>
\\<PREMISES>...</PREMISES>
\\<CONCLUSION>...</CONCLUSION>
\\Classify the conclusion as ``T'' if true, ``F'' if false or ``U'' if unknown based on the premises. Present your answer only between <output></output> tags.
\end{promptbox}

\begin{promptbox}{Scenario 2}
You are an expert logician. You are given a syllogism in FOL with premises between <PREMISES></PREMISES> and conclusion between <CONCLUSION></CONCLUSION> tags.
The FOL BNF grammar to understand and reason in the language is given in the <GRAMMAR></GRAMMAR> tags.
\\<GRAMMAR>...</GRAMMAR>
\\<PREMISES>...</PREMISES>
\\<CONCLUSION>...</CONCLUSION>
\\You are also given the category of the syllogism to help you understand it: Disjunctive.
\\A disjunctive syllogism contains ``$\vee$'' or ``$\oplus$''. Here is an example:
\\<PREMISES>
...</PREMISES>
\\<CONCLUSION>
...</CONCLUSION>
\\Classify the conclusion as ``T'' if true, ``F'' if false or ``U'' if unknown based on the premises. Present your answer only between <output></output> tags.
\end{promptbox}

\section{Experimental Setup}
\label{sec:experimental-setup}

Our experiments extend previous work in~\cite{akl:hal-05248053,akl:hal-05606250} in terms of datasets and setup.

\subsection{Datasets}
\label{subsec:dataset}

To evaluate LM syllogistic reasoning, we consider the  FOLIO~\cite{han2024folio} and P-FOLIO~\cite{han2024p} human-curated syllogism datasets in NL and FOL.
We use CLGC to extend them and create publicly available KR versions  respectively denoted FOLIO-KR\footnote{\url{https://huggingface.co/datasets/HannaAbiAkl/FOLIO-KR}} and P-FOLIO-KR\footnote{\url{https://huggingface.co/datasets/HannaAbiAkl/P-FOLIO-KR}}. 
These extended versions contain the syllogisms generated from FOL into CLIF, CGIF, CLINGO, TFLPLUS, and our custom MINIFOL\textit{x} notations. 
Table~\ref{tab:clgc-example} shows an example of a syllogism from P-FOLIO-KR. 
We also use CLGC to generate the SEF categories for each syllogism in FOLIO and P-FOLIO and include them in our KR extended versions. Table~\ref{tab:dataset-statistics} summarizes label and SEF category statistics for both datasets.

\begin{table}
\caption{CLGC notation example from P-FOLIO-KR.}\label{tab:clgc-example}
\begin{center}
\begin{tabular}{lr}
\toprule
    \multicolumn{1}{c}{Notation} & \multicolumn{1}{c}{Example} \\
\midrule
FOL	& $\forall$x (Has(x, flu) $\rightarrow$ Has(x, influenza)) \\
CLIF	& forall x (has(x, flu) implies has(x, influenza))\\
CLINGO & forall (has(x, flu) -: has(x, influenza)) \\
CGIF & @every *x [(has[(?x  flu)]  has[(?x  influenza)])] \\
MINIFOL2 & all:x (has(x, flu) :- has(x, influenza)) \\
TFLPLUS & -(+H0-+H0) \\
\bottomrule
\end{tabular}
\end{center}
\end{table}

\begin{table}
\caption{Statistics of the extended FOLIO-KR and P-FOLIO-KR datasets.}\label{tab:dataset-statistics}
\begin{center}
\resizebox{\textwidth}{!}{\begin{tabular}{lrrrrrrrrr}
\toprule
    \multirow{2}{*}{Dataset} & \multirow{2}{*}{Total}  & \multicolumn{3}{c}{Label} & \multicolumn{3}{c}{SEF} \\
    \cmidrule(r{0.2em}l{0.2em}){3-5} \cmidrule(r{0.2em}l{0.2em}){6-9}
    & & True & False & Unknown & Categorical & Hypothetical & Disjunctive & Complex \\
\midrule
FOLIO-KR & 1204	& 460 & 351 & 393 & 17 & 67 & 653 & 467 \\
P-FOLIO-KR & 301 & 122 & 69 & 110 & 9 & 13 & 40 & 239\\
\bottomrule
\end{tabular}}
\end{center}
\end{table}

\subsection{Protocol}
\label{subsec:experiments}

We treat the syllogistic reasoning problem as a multi-class classification task with the values \{True, False, Unknown\}. 
Experiments were performed on SLMs in SFT and ZS settings on P-FOLIO-KR and FOLIO-KR.
SFT experiments were performed on A100 GPUs and ZS experiments on L4 GPUs.

\subsubsection{Supervised Fine-Tuning.}
\label{subsubsec:sft-experiments}

In SFT, the dataset is split into the Train, Val and Test sets. 
In the Train and Val sets, each syllogism (i.e. premises and conclusion) in a notation is associated to its truth label.
In the Test set, the truth labels are omitted and the model predicts them for each syllogism during inference. The FOLIO-KR and P-FOLIO-KR datasets are split into stratified frozen Train, Val, and Test sets for reproducibility. Stratification balances the splits and ensures all sets have similar truth label and SEF category distributions. 
Table~\ref{tab:dataset-split-statistics} shows the statistics for the labels and SEF categories in the frozen splits.

We build on previous SLM results~\cite{han2024folio,han2024p} by limiting our selection to Flan-T5-small\footnote{\url{https://huggingface.co/google/flan-t5-small}} and Flan-T5-large\footnote{\url{https://huggingface.co/google/flan-t5-large}} to study notation-based reasoning on smaller, frugal language models as defined in~\cite{wang2025comprehensive}. The Flan-T5 encoder-decoder architecture makes the model a good candidate for both understanding and generating text which is required for syllogistic reasoning. Furthermore, Flan-T5 has already exhibited good performances on this task~\cite{han2024folio,han2024p} with respect to other similar models in size and architecture (e.g. RoBERTa). Training is performed on 5 epochs with a batch size of 4. The models are initialized with a seed to ensure reproducibility. We list all training parameters on GitHub\footnote{\url{https://github.com/HannaAbiAkl/clgc/tree/main/experiments/folio/notebooks}}.

\begin{table}
\caption{Frozen split statistics.}\label{tab:dataset-split-statistics}
\begin{center}
\resizebox{\textwidth}{!}{\begin{tabular}{llrrrrrrrr}
\toprule
    \multicolumn{1}{c}{\multirow{2}{*}{Dataset}} & \multicolumn{1}{c}{\multirow{2}{*}{Split}} & \multicolumn{1}{c}{\multirow{2}{*}{Total}} & \multicolumn{3}{c}{Label} & \multicolumn{4}{c}{SEF} \\
    \cmidrule(r{0.2em}l{0.2em}){4-6} \cmidrule(r{0.2em}l{0.2em}){7-10} 
    & & & True & False & Unknown & Categorical & Hypothetical & Disjunctive & Complex \\
\midrule
\multirow{ 3}{*}{P-FOLIO-KR} & Train & 144	& 58 & 33 & 53 & 3 & 4 & 21 & 116 \\
 & Val & 96	& 39 & 22 & 35 & 6 & 6 & 15 & 69 \\
 & Test & 61 & 25 & 14 & 22 & 0 & 3 & 4 & 54 \\
 \hline
\multirow{ 3}{*}{FOLIO-KR} & Train & 800 & 316 & 225 & 259 & 11 & 43 & 432 & 314 \\
 & Val & 201	& 72 & 64 & 65 & 3 & 13 & 115 & 70 \\
 & Test & 203 & 72 & 62 & 69 & 3 & 11 & 106 & 83 \\
\bottomrule
\end{tabular}}
\end{center}
\end{table}

\subsubsection{Zero-Shot.}
\label{subsubsec:zs-experiments}

In Zero-Shot, the model is given the syllogism and a prompt instructing it to use logical reasoning to predict the truth value from the possible labels. For these experiments, we also restrict ourselves to small, decoder-based models ideal for text generation (i.e. less than 10 billion parameters~\cite{wang2025comprehensive}). Our selection comprises Gemma-2-2b-it\footnote{\url{https://huggingface.co/google/gemma-2-2b-it}}, Llama-3.2-3b-instruct\footnote{\url{https://huggingface.co/meta-llama/Llama-3.2-3B-Instruct}}, and Phi-3.5-mini-instruct\footnote{\url{https://huggingface.co/microsoft/Phi-3.5-mini-instruct}}. Each model has its particularity aside from being pre-trained on different corpora: Gemma is an all-around general model, Llama is open-source, and Phi is specialized in reasoning and mathematical tasks. All models are trained on Scenarios 1 and 2 with the prompts of Section~\ref{subsec:lm-reasoning-evaluation}.

\section{Results and Discussion}
\label{sec:results}

Since our datasets are unbalanced in truth labels (Table \ref{tab:dataset-statistics}), we measure and report model performance using the F1 score in our SFT and ZS experiments. In ZS, we also compute the Absolute Gain (AG) to measure the change in F1 between Scenario 1 (S1) and 2 (S2) prompts: $AG_{F1} = F1_{\text{S2}} - F1_{\text{S1}}$.

\subsection{Supervised Fine-Tuning}
\label{subsec:sft-results}

\subsubsection{Notation performance depends on model and dataset.} In Table~\ref{tab:results-pfolio-sft-small}, we observe that for P-FOLIO-KR, Flan-T5-small performs best on abstract notations (e.g. TFLPLUS) and combinations thereof (e.g. CLIF + TFLPLUS, NL + CLIF + TFLPLUS). 
TFLPLUS performs almost 3 times better than NL with only limited SFT training (144 and 96 syllogisms respectively in Train and Val sets) compared to the model pre-training on NL.
FOL and FOL-like notations (i.e. MINIFOL2, CLINGO) perform midway between NL and TFLPLUS. 
Of all notations, TFLPLUS has the simplest syntax, with $+$ and $-$ tokens already seen by Flan-T5-small's tokenizer during pre-training.
We can hypothesize that this is the reason why the model learns the notation well as opposed to MINIFOL2 for example.
Flan-T5-large's performance shows a different trend, where NL is the best notation followed by CLIF which is a more compact version of NL. The model still performs better on TFLPLUS than on FOL-like notations. We can hypothesize that the additional NL data seen by Flan-T5-large in pre-training compared to Flan-T5-small is the reason for the boost in performance in NL, and that some symbols in FOL and FOL-like notations (e.g. quantifiers) are harder to learn, due to their under-representation in the pre-training dataset. 

For FOLIO-KR, we observe that Flan-T5-small performs best on a combination of NL and CLIF. FOL and FOL-like notations (CLINGO, CGIF) rank midway in performance while TFLPLUS is among the worst performers. 
FOLIO-KR's larger Train and Val splits (800 and 201 syllogisms respectively) compared to P-FOLIO-KR could cause this boost in performance of notations with well-seen tokens like NL, CLIF and NL + CLIF. 
This increase did not help notations like TFLPLUS and MINIFOL which rely heavily on potentially less seen symbolic tokens (i.e. algebraic and boolean operators). This might also explain the relative good performance of $\text{MINIFOL}_x$, which eliminate symbols compared with MINIFOL.
For Flan-T5-large, we observe results on FOLIO-KR similar to those of Flan-T5-small. NL is the best notation, followed by CLIF. NL + CLIF ranks between the two notations. MINIFOL2 and CLINGO, which share similar abstraction and complexity (e.g. shared symbols like \textit{:-}) rank close behind. We can hypothesize that FOL and CGIF --the most complex notations (i.e. containing the most symbols)-- and TFLPLUS --the most abstract one (i.e. everything is represented with $+$ and $-$ tokens)-- are the hardest to learn at scale.

\begin{table}
	\caption{Flan-T5 SFT results. Best results are in bold, second-best are underlined.
 }\label{tab:results-pfolio-sft-small}
	\centering

	\subfloat[P-FOLIO-KR]{
		\resizebox{0.35\textwidth}{!}{\begin{tabular}{llr}
			\toprule
			\multicolumn{1}{c}{Model} & \multicolumn{1}{c}{Notation} & \multicolumn{1}{c}{F1} \\
			\midrule
			\multirow{ 10}{*}{Small} & CLIF + TFLPLUS & \textbf{0.367} \\
		 	& TFLPLUS &	\underline{0.350} \\
		 	& NL + CLIF + TFLPLUS	&	0.339 \\
            & CLIF	& 0.298 \\
		 	& CGIF	& 0.295 \\
            & FOL	& 0.289 \\
		 	& CLINGO	& 0.276 \\
		 	& MINIFOL2	& 0.239 \\
            & NL + CLIF	& 0.187 \\
		 	& NL	&	0.126 \\
		 	\midrule
		 	\multirow{ 9}{*}{Large} & NL	&	\textbf{0.688}\\
		 	& CLIF	& \underline{0.510}\\
		 	& NL + CLIF + TFLPLUS	& 0.508\\
		 	& NL + CLIF	& 0.490\\
		 	& TFLPLUS	&	0.460\\
		 	& CGIF	&	0.368\\
		 	& MINIFOL2	& 0.324\\
		 	& FOL	&	0.278\\
		 	& CLINGO	& 0.272\\
			\bottomrule
		\end{tabular}}
	}
    \hspace{5em}
	\subfloat[FOLIO-KR]{
		\resizebox{0.23\textwidth}{!}{\begin{tabular}{llr}
			\toprule
			\multicolumn{1}{c}{Model} & \multicolumn{1}{c}{Notation} & \multicolumn{1}{c}{F1} \\
			\midrule
			\multirow{ 12}{*}{Small}  & NL + CLIF	& \textbf{0.447}\\
		 	& NL	& \underline{0.435}\\
		 	& CLIF	& 0.410\\
		 	& NL + FOL	& 0.395\\
		 	& FOL	& 0.382\\
		 	& MINIFOL2	& 0.362\\
		 	& MINIFOL3	& 0.357\\
		 	& MINIFOL4	& 0.356\\
		 	& CLINGO	& 0.353\\
		 	& CGIF	& 0.339\\
		 	& TFLPLUS	& 0.291\\
		 	& MINIFOL	& 0.190\\
			\midrule
			\multirow{ 9}{*}{Large} & NL	& \textbf{0.658}\\
		 	& NL + CLIF	& \underline{0.638}\\
		 	& CLIF	& 0.614\\
		 	& CLINGO	& 0.601\\
		 	& MINIFOL2	& 0.581\\
		 	& FOL	& 0.506\\
		 	& TFLPLUS	& 0.534\\
		 	& CGIF	& 0.181\\
		 	& MINIFOL	& 0.177\\
			\bottomrule
		\end{tabular}}
	}
\end{table}

\subsubsection{Performance increases with model size independently of abstraction.}
\label{subsubsec:sft-analysis-notation-trends}

\begin{figure}[t!]
    \centering
    \subfloat[P-FOLIO-KR.]{
        \includegraphics[width=0.45\textwidth]{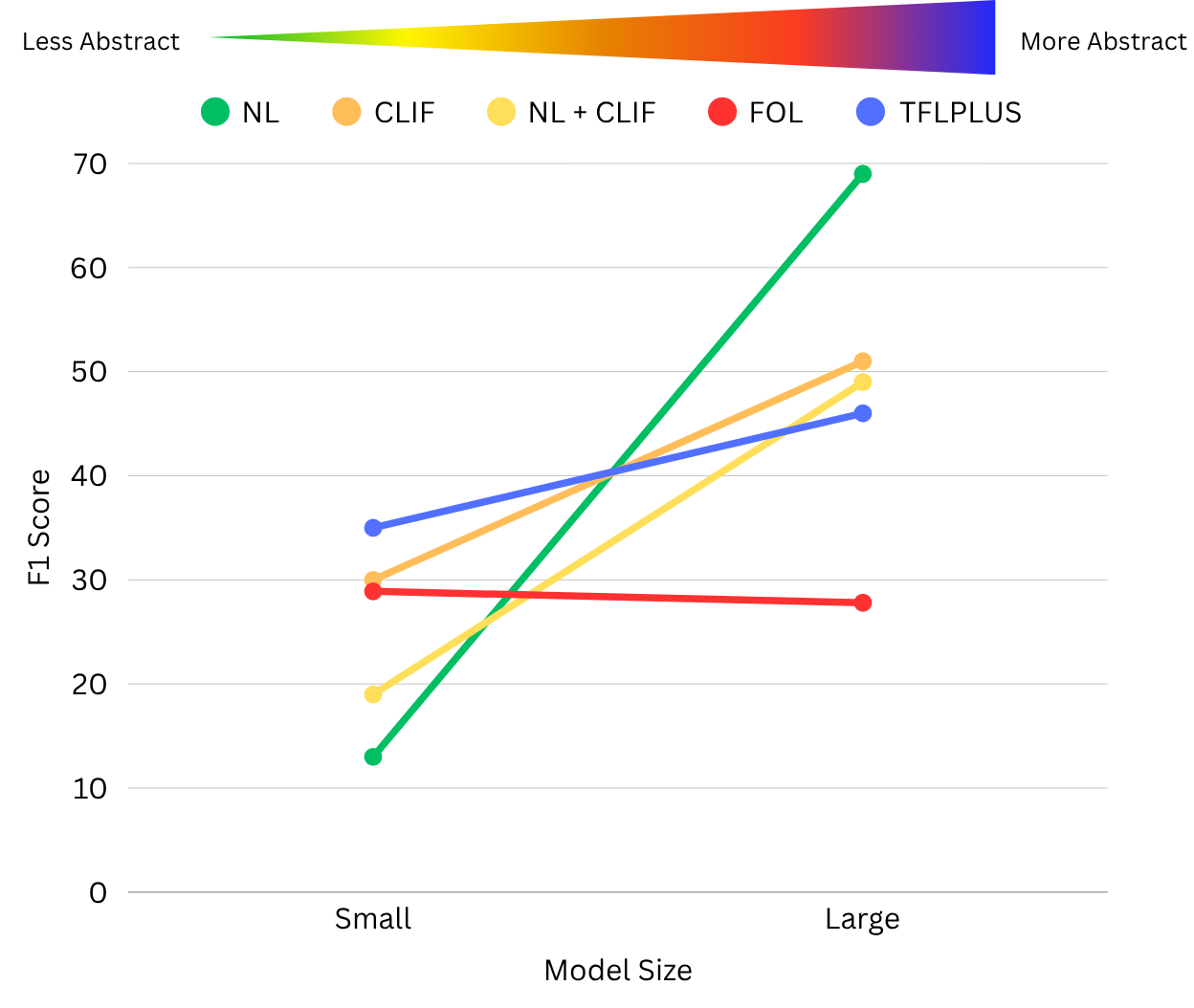}
    }
    \hfill 
    \subfloat[FOLIO-KR.]{
        \includegraphics[width=0.45\textwidth]{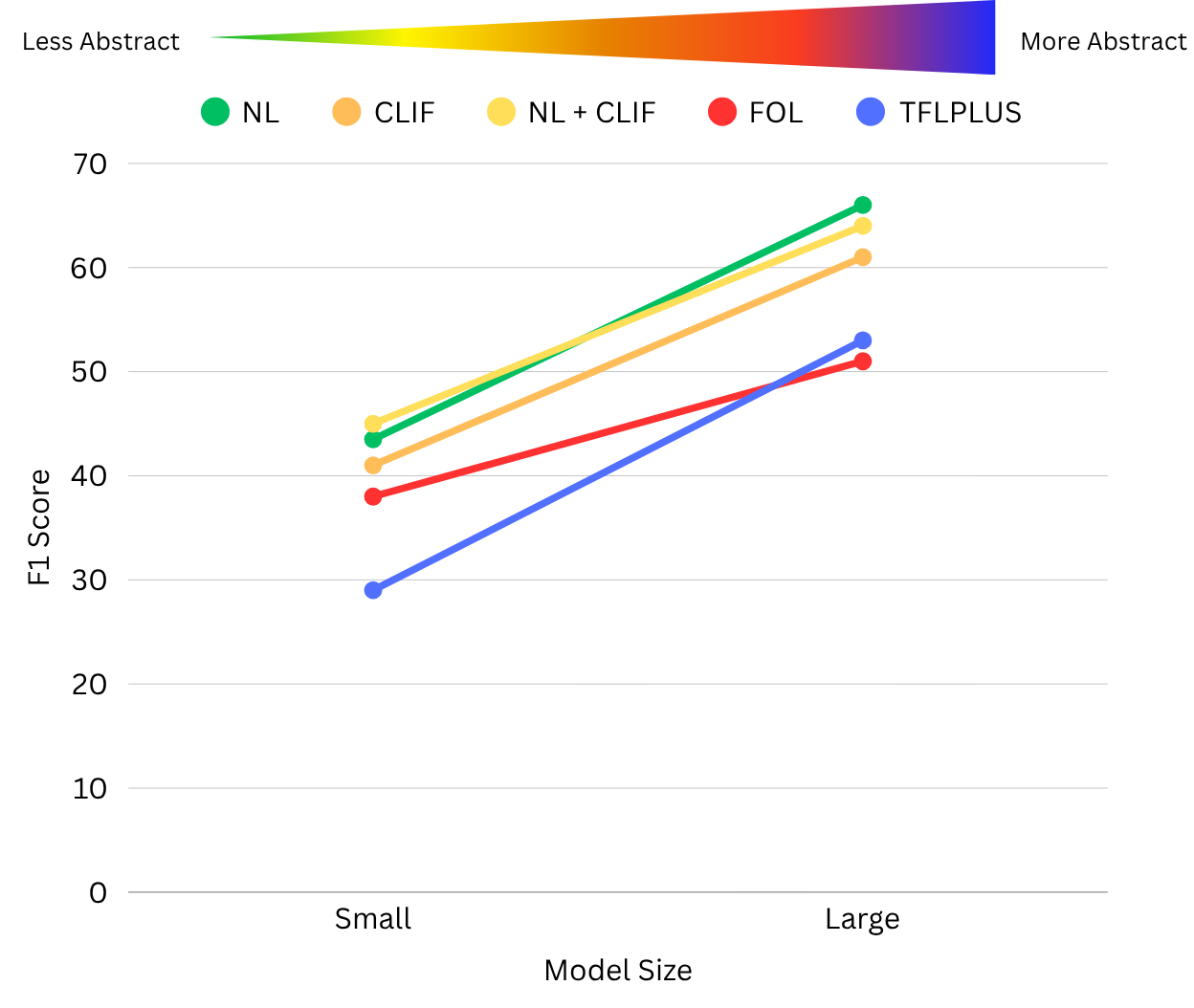}
    }
    \caption{Model size impact on notation performance for Flan-T5.}
\label{fig:folio-sft-notation-impact}
\end{figure}

Figure \ref{fig:folio-sft-notation-impact} shows the performance trend of different notations based on model size on P-FOLIO-KR and FOLIO-KR in SFT. On P-FOLIO-KR, we observe that all notations, except FOL, exhibit an upward trend when scaling Flan-T5. We also observe that Flan-T5-small performs better on more abstract notations but the trend changes when the model is scaled.
Similarly, on FOLIO-KR, performance increases for all notations  with scaling, with FOL experiencing a slight lesser growth.
We can hypothesize that the behavior of FOL is due to the complexity of its syntax in terms of tokens compared to more abstract but syntactically simpler notations (i.e. TFLPLUS). This may also explain FOL's downward trend on P-FOLIO-KR since P-FOLIO-KR's syllogisms are designed to be longer and more complex in reasoning depth than FOLIO-KR's~\cite{han2024p}.

\subsubsection{Combining natural and symbolic notations yields more conservative reasoning.}
\label{subsubsec:sft-analysis-constrained-reasoning}

\begin{figure}[t!]
    \centering
    \subfloat[True label.]{
        \includegraphics[width=0.3\textwidth]{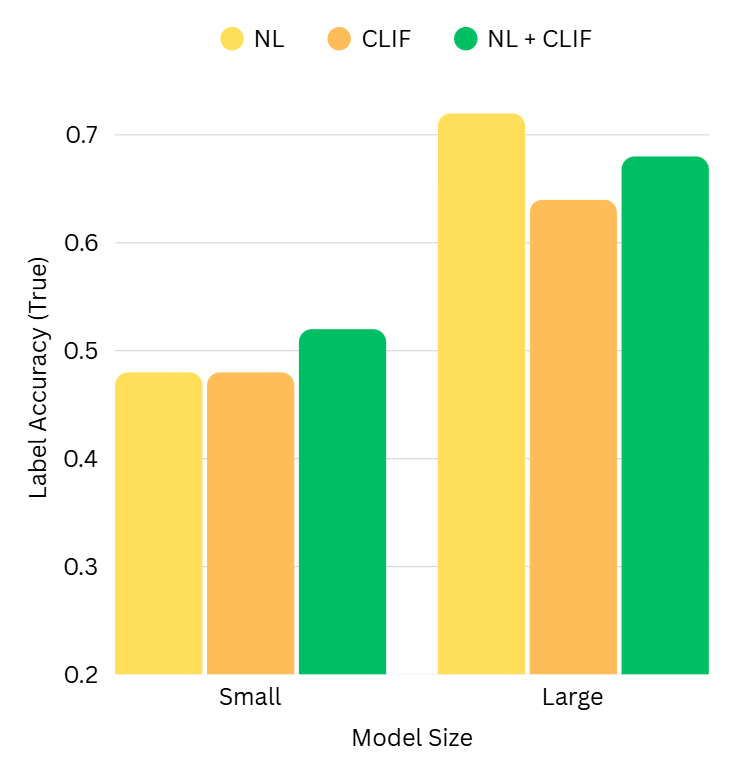}
    }
    \hfill 
    \subfloat[False label.]{
        \includegraphics[width=0.3\textwidth]{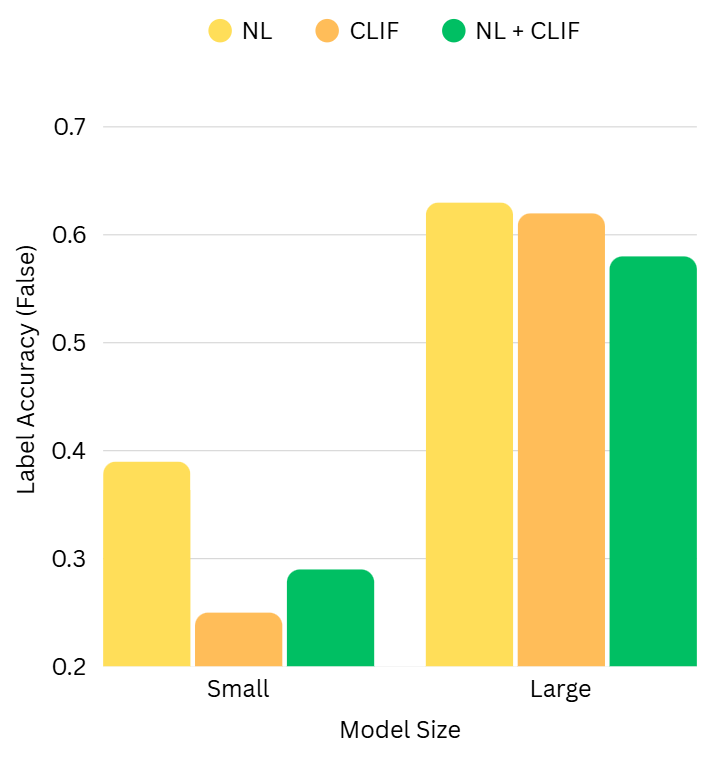}
    }
    \hfill
    \subfloat[Unknown label.]{
        \includegraphics[width=0.3\textwidth]{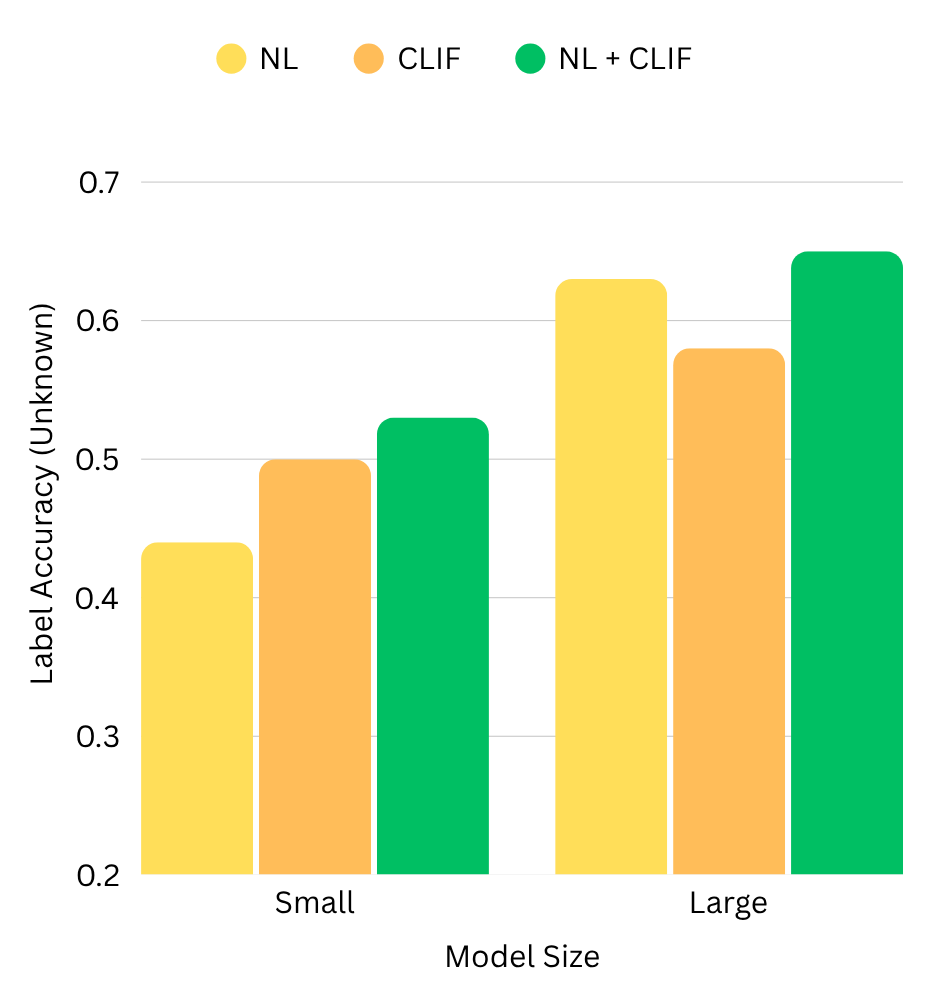}
    }
    \caption{Flan-T5-small and large NL, CLIF and NL + CLIF per-label accuracy on FOLIO-KR SFT.}
\label{fig:folio-sft-label-comparison}
\end{figure}

Figure \ref{fig:folio-sft-label-comparison} compares the per-label accuracy of the best notations (i.e. NL, CLIF and NL + CLIF) with Flan-T5 small and large models on FOLIO-KR. For the label True, we observe that CLIF is the worst performer in the small and large models. Results also show that while NL + CLIF performs better than NL in Flan-T5-small, it does not outperform it in the large model. 
We hypothesize that CLIF makes the model reasoning more conservative by limiting True predictions. This is further substantiated by the results on False and Unknown.
For the False label, CLIF benefits from the biggest increase in performance when scaling from small to large and outperforms NL + CLIF in Flan-T5-large. One hypothesis is that the increase in performance on the False label is related to the compactness of CLIF compared to NL which makes it favor reasoning by refutation (i.e. predicting False). 
Results on the Unknown label show that NL + CLIF bests the other notations at small and large scales. We observe that at small scale, CLIF performs better than NL. We can hypothesize that combining a compact notation like CLIF with NL improves model reasoning in uncertainty. This may be seen as the ``hardest'' form of reasoning since it cannot validate (i.e. True) or refute (i.e. False) based on the premises and conclusion at hand. It can be argued that this type of reasoning requires more abstraction which NL + CLIF allows. 
In view of the general tendency of LMs to hallucinate instead of stating they do not know the answer, the global conservative behavior of CLIF (or NL + CLIF) could be seen as a potential remediation, where predicting Unknown could also be seen as the best compromise.

\subsubsection{NL + CLIF as a conservative SFT reasoning language representation.}
\label{subsubsec:sft-analysis-label-prediction-distribution}

\begin{figure}[t!]
\captionsetup[subfigure]{justification=centering}
    \centering
    \subfloat[True label.]{
        \includegraphics[width=0.9\textwidth]{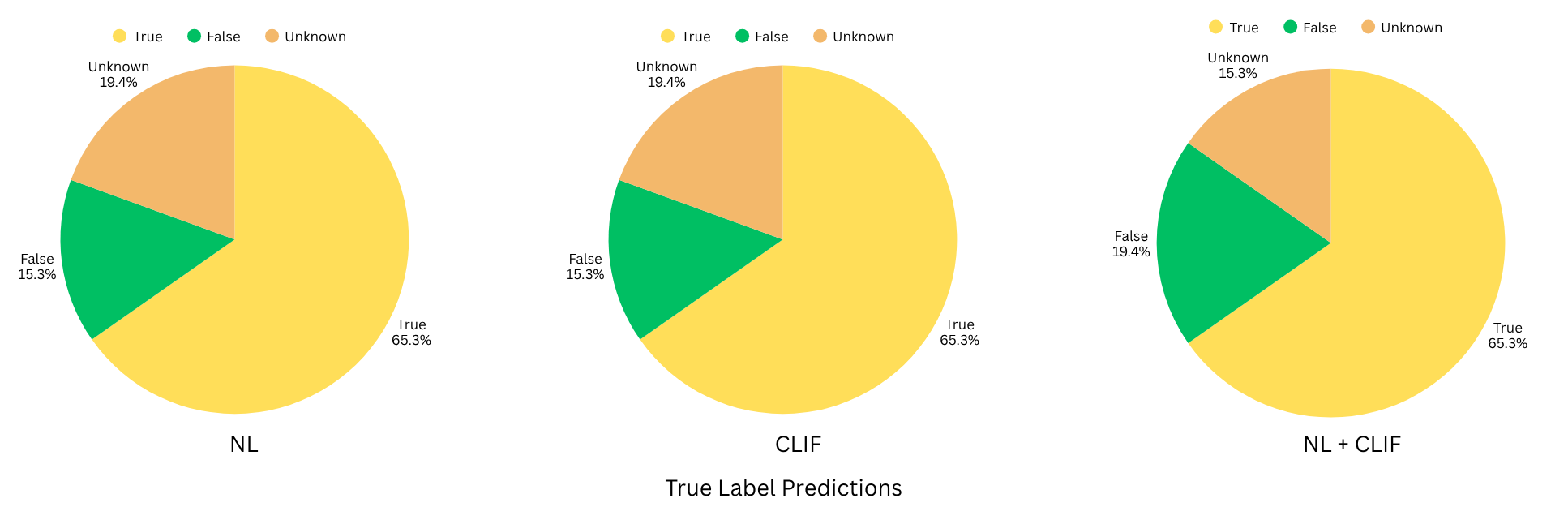}
    }
    
    \subfloat[False label.]{
        \includegraphics[width=0.9\textwidth]{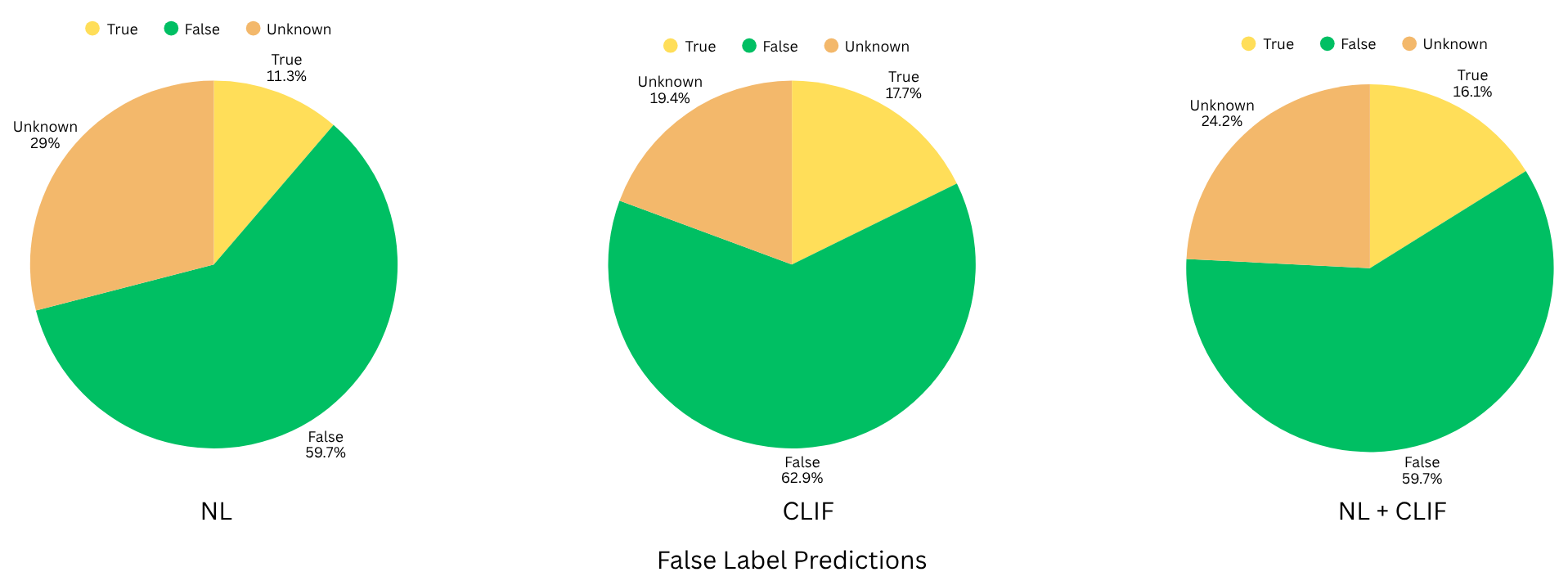}
    }
    
    \subfloat[Unknown label.]{
        \includegraphics[width=0.9\textwidth]{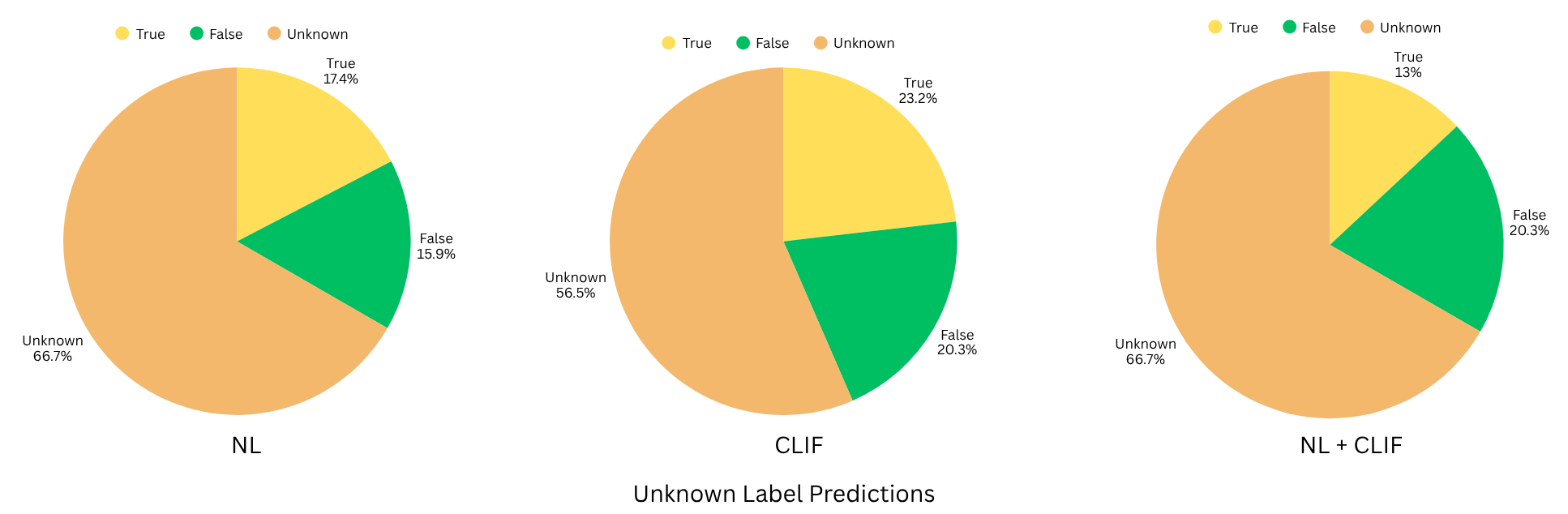}
    }
    \caption{Flan-T5-large NL, CLIF and NL + CLIF label prediction distribution on FOLIO-KR SFT.}
\label{fig:folio-sft-label-prediction-distribution-true}
\end{figure}

In Figure~\ref{fig:folio-sft-label-prediction-distribution-true}, we observe that combining NL and CLIF reverses the trend of misclassification on the True label. NL + CLIF favors False as a mistake contrary to NL and CLIF. 
For the False label, we observe that NL + CLIF is more balanced in misclassifying False as either True or Unknown compared to NL, with CLIF alone being the most balanced notation. 
Results also show that CLIF is the best at classifying False, which strengthens our claim that this compact notation improves a specific type of reasoning (i.e. reasoning by refutation) compared to NL. 
For the Unknown label, we observe that NL and NL + CLIF are tied in performance but CLIF and NL + CLIF are more likely to misclassify the label as False than NL.
Since label distribution in FOLIO-KR largely favors True (Table \ref{tab:dataset-split-statistics}), we hypothesize that introducing a more compact notation (i.e. CLIF) and combinations (i.e. NL + CLIF) alters model reasoning. 
Figure~\ref{fig:folio-sft-label-comparison} supports this hypothesis as scaling the model shows that NL is the best in predicting True, CLIF alone bests NL + CLIF on False, and NL + CLIF is consistently the best at predicting Unknown.
This is a promising avenue for combining KR notations to refine model reasoning, possibly in a multi-stage neuro-symbolic pipeline with NL as a first step and refinement in CLIF or NL + CLIF. 

\subsection{Zero-Shot}
\label{subsec:zs-results}

\begin{table}
	\caption{ZS results for S1 and S2. Best results are in bold, second-best are underlined.}
	\label{tab:results-pfolio-zs-gemma}
	\begin{center}
		\resizebox{0.8\textwidth}{!}{\begin{tabular}{lllrrr}
			\toprule
			\multicolumn{1}{c}{Dataset} & \multicolumn{1}{c}{Model} & \multicolumn{1}{c}{Notation} & \multicolumn{1}{c}{$F1_{\text{S1}}$} & \multicolumn{1}{c}{$F1_{\text{S2}}$} & \multicolumn{1}{c}{AG} \\
			\midrule
			\multirow{22}{*}{\textbf{P-FOLIO-KR}} & \multirow{ 8}{*}{Gemma-2-2b-it} & NL	& 0.353 & \textbf{0.464} & \textbf{+0.111}\\
			& & FOL	& 0.302 & \underline{0.400} & \underline{+0.098}\\
			& & CLIF & 0.302 & 0.368 & +0.066\\
			& & CGIF	& 0.373 & 0.221 & -0.152\\
			& & CLINGO	& \underline{0.401} & 0.343 & -0.058\\
			& & MINIFOL2 & 0.295 & 0.364 & +0.069\\
			& & TFLPLUS	& 0.298 & 0.247 & -0.051\\
			& & NL + CLIF	& \textbf{0.403} & 0.375 & -0.048\\
			\cmidrule{2-6}
			& \multirow{ 7}{*}{Llama-3.2-3b-instruct} & MINIFOL2	& \underline{0.041} & \textbf{0.049} & +0.008\\
			& & FOL	& 0.021 & \underline{0.048} & \underline{+0.027}\\
			& & CLIF	& 0.027 & 0.020 & -0.007\\
			& & NL	& 0.035 & 0.038 & +0.003\\
			& & CGIF	& \textbf{0.045} & 0.043 & -0.002\\
			& & CLINGO	& 0.017 & 0.045 & \textbf{+0.028}\\
			& & TFLPLUS	& 0.016 & 0.014 & -0.002\\
			\cmidrule{2-6}
			& \multirow{ 7}{*}{Phi-3.5-mini-instruct} & CLIF	& \textbf{0.206} & 0.141 & -0.065\\
			& & NL	& 0.101 & \underline{0.143} & \underline{+0.042}\\
			& & FOL	& \underline{0.163} & 0.123 & -0.040\\
			& & CGIF	& 0.059 & 0.066 & +0.007\\
			& & CLINGO	& 0.077 & \textbf{0.186} & \textbf{+0.109}\\
			& & MINIFOL2	& 0.129 & 0.042 & -0.087\\
			& & TFLPLUS	& 0.079 & 0.045 & -0.034\\
			\midrule
			\multirow{ 8}{*}{\textbf{FOLIO-KR}} & \multirow{ 8}{*}{Gemma-2-2b-it} & NL + CLIF	& 0.234	& \textbf{0.316} & \underline{+0.082} \\
			& & NL	& 0.224	& \underline{0.311} &	\textbf{+0.087} \\
			& & CLIF	& \underline{0.266}	& 0.272 &	+0.006 \\
			& & CLINGO	& \textbf{0.270}	& 0.265 &	-0.005 \\
			& & CGIF	&	0.178 & 0.208 &	+0.030 \\
			& & FOL	& 0.210 & 0.137 &	-0.073	\\
			& & MINIFOL2	& 0.193 & 0.258 &	+0.065	\\
			& & TFLPLUS	& 0.184 & 0.184	& 0 \\
			\bottomrule
		\end{tabular}}
	\end{center}
\end{table}

\subsubsection{The choice of KR notation depends on model and training paradigm.}
\label{subsubsec:zs-analysis-different-notation-paradigms}

Comparing Tables \ref{tab:results-pfolio-sft-small} and \ref{tab:results-pfolio-zs-gemma} shows variations in notation performance with model and training paradigm. In SFT, NL, CLIF and the combination thereof (i.e. NL + CLIF) seem to be the best-suited for models like Flan-T5. 
In ZS, there is no prior fine-tuning phase on the notation before inference and model performance varies based on the notation abstraction and the model's initial pre-training data.
To the best of our knowledge, Phi is trained on synthetic and logical data (e.g. mathematical) which might make it more suited for reasoning in abstract notations (e.g. CLINGO) with SEF descriptions. 
Gemma is, to the best of our knowledge, mostly pre-trained on text data which fits the hypothesis that it should perform better with NL and NL-like (i.e. CLIF) notations with SEF descriptions in those notations. 
One possible explanation for Llama's performance on FOL-like notations may be its distilled pre-training data which is highly filtered content from the web, making it less averse to notations with symbolic tokens than models like Gemma.  
These observations strengthen our claim in Section~\ref{subsec:sft-results} regarding multi-stage model design based on notation and learning paradigm.

\subsubsection{SEF description impact varies with model and notation.}
\label{subsubsec:zs-analysis-sef-category-impact}

Table \ref{tab:results-pfolio-zs-gemma} shows the ZS results for scenarios S1 and S2 on P-FOLIO-KR and FOLIO-KR. For Gemma on P-FOLIO-KR, we observe that NL gets the biggest AG increase in S2. The performance of NL + CLIF, which is the best by default, decreases when adding descriptions. S2 also lowers the performance of abstract notations (i.e. CLINGO, TFLPLUS). With no fine-tuning in ZS, we can hypothesize that the impact of descriptions varies with model sensitivity to the notation. 
Since Gemma is mainly pre-trained on text data, new notations or combinations (e.g. TFLPLUS, NL + CLIF) might lower model performance when prompted with added descriptions.
For Llama, its pre-training on distilled data from highly filtered web content makes it less averse to symbolic tokens than models like Gemma. 
This is shown by its performance on FOL and FOL-like notations (e.g. MINIFOL2, CLINGO) in S1 and S2. 
Finally, Phi being pre-trained on synthetic and logical (e.g. mathematical) data may explain its performance increase when adding descriptions for abstract notations.

For FOLIO-KR, previous experimental results~\cite{akl:hal-05248053} have shown that Gemma bests Llama and Phi in ZS setting, making it the sole focus of S2 experiments for this dataset. Results strengthen the observations seen in P-FOLIO as reasoning improves when adding descriptions on NL-like notations like CLIF and the combination NL + CLIF. TFLPLUS is a curious case that receives 0 AG in S2 on this larger dataset. One possible explanation for this behavior might be that the model plateaued out due to the high abstraction level of the notation. 

{
\setlength{\tabcolsep}{6pt}
\begin{table}
\begin{center}
\caption{Gemma-2-2b-it ZS Scenario 2 runtime (in minutes) on FOLIO-KR. Best results in bold, second-best underlined.}\label{tab:results-folio-zs-gemma}
\resizebox{\textwidth}{!}{\begin{tabular}{lrrrrrrrr}
\toprule
Notation & \multicolumn{1}{c}{NL + CLIF} & \multicolumn{1}{c}{NL} & \multicolumn{1}{c}{CLIF} & \multicolumn{1}{c}{CLINGO} & \multicolumn{1}{c}{CGIF} & \multicolumn{1}{c}{FOL} & \multicolumn{1}{c}{MINIFOL2} & \multicolumn{1}{c}{TFLPLUS} \\
\midrule
Runtime (min) & 37 & 30 & \underline{23} & 25 & 27 & 28 & 24 & \textbf{21} \\
\bottomrule
\end{tabular}}
\end{center}
\end{table}
}

\subsubsection{KR notations show faster reasoning inference.}
\label{subsubsec:zs-analysis-runtime}

Table \ref{tab:results-folio-zs-gemma} shows the ZS inference runtime of different notations with Gemma-2 in S2. The results are limited to Gemma since it is the overall best-performing model and to FOLIO-KR since it is 4 times larger than P-FOLIO-KR, making runtime differences more pronounced. Observations show that NL has the longest runtime, making it the most computationally expensive notation. TFLPLUS is the most compact and fastest syntax. CLIF appears as a good compromise between runtime and performance based on Table \ref{tab:results-pfolio-zs-gemma}. Combining notations (e.g. NL + CLIF) can increase performance in ZS at the expense of longer runtimes. These results further support our claim that notation choice depends on multiple factors and should be considered in the design of multi-stage reasoning pipelines. 

\section{Conclusion}
\label{sec:conclusion}

In this work, we introduced the CLGC framework, which converts syllogisms represented in FOL into alternative formal notations, to study of the impact of KR notations on syllogistic reasoning in SLMs.
Our findings show that model size increase performance independently of notation abstraction. 
Additionally, combinations of notations like NL + CLIF shift reasoning on syllogisms, improving refutation and uncertainty.
Our results also show that providing descriptions of syllogism categories in  prompt can improve model performance, but the improvement varies depending on the notation and model used. 
We released the FOLIO-KR and P-FOLIO-KR datasets and made the CLGC framework and experiments publicly available. 
CLGC demonstrates that small, frugal models can perform well on small reasoning datasets with the right choice of notation. 
The dichotomy between natural and abstract notations paves the way for future work on neuro-symbolic model architectures combining both to carry out complex, multi-stage reasoning in different training paradigms. 

\begin{credits}
\subsubsection{\ackname}
This work was supported by the French government through the France 2030 investment plan managed by the National Research Agency (ANR), as part of the Initiative of Excellence Université Côte d’Azur (ANR-15-IDEX-01), by the 3IA Côte d’Azur (ANR-19-P3IA-0002), by the Université Côte d’Azur’s Center for High-Performance Computing and by the Data ScienceTech Institute.

\subsubsection{\discintname}
The authors have no competing interests to declare that are
relevant to the content of this article.
\end{credits}
%
%
%
%
\newpage
\bibliographystyle{splncs04}
\bibliography{bibliography}

\end{document}